\documentclass{article}

\PassOptionsToPackage{numbers, compress}{natbib}
\usepackage[main, preprint]{neurips_2026}

\usepackage[utf8]{inputenc}
\usepackage[T1]{fontenc}
\usepackage{hyperref}
\usepackage{url}
\usepackage{booktabs}
\usepackage{amsfonts}
\usepackage{amssymb}
\usepackage{amsmath}
\usepackage{nicefrac}
\usepackage{microtype}
\usepackage{xcolor}
\usepackage{graphicx}
\usepackage{multirow}
\usepackage{siunitx}
\usepackage{subcaption}
\usepackage{tikz}
\usepackage{pgfplots}
\usepackage{pgfplotstable}
\usepgfplotslibrary{groupplots}
\usepackage{pgf-pie}
\usetikzlibrary{matrix}
\pgfplotsset{compat=1.18}

\definecolor{darkblue}{RGB}{0, 51, 102}
\definecolor{darkred}{RGB}{153, 0, 0}
\definecolor{darkgreen}{RGB}{0, 102, 51}

\definecolor{primaryblue}{RGB}{24, 86, 178}
\definecolor{secondaryblue}{RGB}{35, 120, 195}

\definecolor{plum}{RGB}{120, 28, 110}
\definecolor{forest}{RGB}{34, 139, 34}
\definecolor{ocean}{RGB}{0, 119, 190}

\hypersetup{
    colorlinks=true,
    linkcolor=plum,
    citecolor=forest,
    filecolor=ocean,
    urlcolor=ocean,
    pdftitle={GRASP: Guided Residual Adapters with Sample-wise Partitioning},
}

\title{GRASP: Guided Residual Adapters with Sample-wise Partitioning}

\author{%
  Felix Nützel\textsuperscript{1} \quad
  Mischa Dombrowski\textsuperscript{1} \quad
  Bernhard Kainz\textsuperscript{1,2}\\
  \textsuperscript{1} Friedrich-Alexander-Universität Erlangen-Nürnberg, GER\\
  \textsuperscript{2} Imperial College London, UK\\
  \texttt{felix.nuetzel@fau.de}
}

\begin{document}

\maketitle

\begin{abstract}
Text-to-image flow matching transformers degrade sharply in long-tail settings:
tail-class outputs collapse in fidelity and diversity, limiting their value as synthetic augmentation for rare conditions.
We trace this to low head-versus-tail gradient alignment during fine-tuning, an optimization-level pathology that
conditioning- and sampling-side interventions do not address.
We propose GRASP (Guided Residual Adapters with Sample-wise Partitioning): a deterministic partition of the conditioning space,
paired with group-specific residual adapters in the transformer feedforward layers, that leaves the flow-matching objective
and the sampler untouched.
In conditional flow matching, condition values index distinct sets of probability paths, which makes the conditioning a structurally natural axis along which to factorize residual updates.
Because the partition is static, every tail sample is guaranteed to update its assigned expert, which bypasses extreme long-tail failure modes. GRASP is non-invasive and composable: on MIMIC-CXR-LT, combining GRASP with self-guided minority sampling at inference outperforms its learned counterpart on 10 of 12 metrics.
GRASP itself reduces overall Fréchet Inception Distance (FID) by up to 80\% and lifts tail-class coverage by up to 44\% over full fine-tuning,
learned-routing mixture-of-experts and long-tail baselines. Performance is stable across random reassignments of labels to experts and across expert counts, as long as no expert is starved of samples.
Used as training data for a downstream DenseNet classifier on NIH-CXR-LT, GRASP synthetics significantly outperform every non-GRASP alternative on macro F1 and yield nonzero F1 on 9 of 13 classes versus 3 of 13 from real data, while maintaining comparable macro F1.
Results on ImageNet-LT confirm the mechanism is not tied to medical inductive bias. Code is available at \url{https://github.com/Felix-012/GRASP-LT}
\end{abstract}

\section{Introduction}
\label{sec:intro}

Text-to-image flow matching models have transformed visual generation by enabling controllable and photorealistic synthesis at scale.
Yet, their performance degrades markedly in long-tail distributions, where rare classes dominate real-world scenarios, like medical imaging.
In such settings, models prioritize frequent head classes, leading to mode collapse in tail classes: generated images for rare pathologies exhibit poor fidelity and severely limited diversity.
The result is a systematic collapse of diversity in rare conditions, which limits the potential of synthetic data augmentation and undermines downstream robustness. This hampers critical applications, such as augmenting datasets for underrepresented diseases, where diverse, high-quality synthetics are essential for robust training.
\begin{figure}[t]
    \centering
    \includegraphics[width=\linewidth, clip]{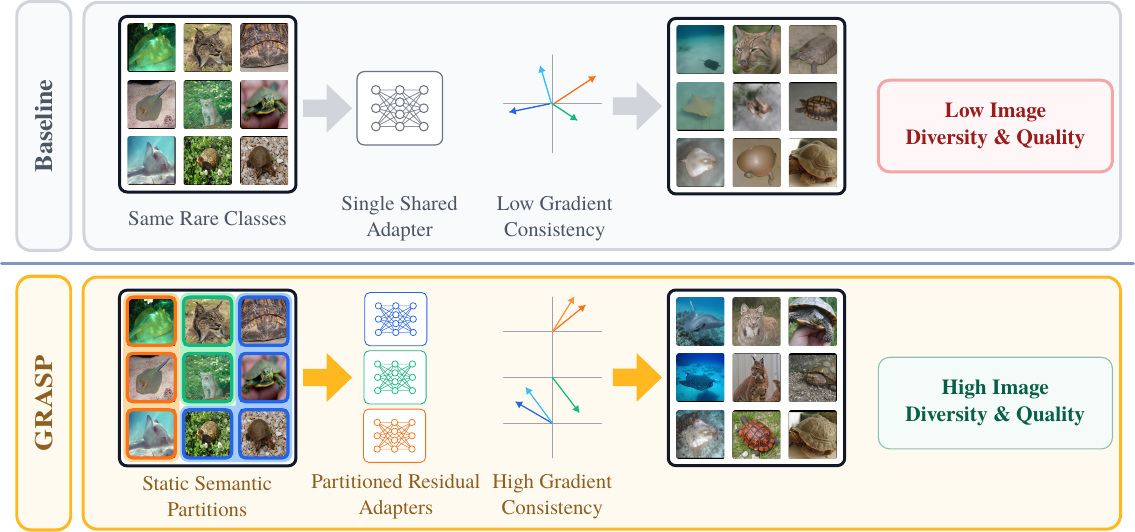}
    \caption{Our proposed method uses parallel GRASP adapters to gain more aligned gradient updates when training a diffusion transformer. This enables the model to generate images of higher quality and diversity, in particular for rare classes.}
    \label{fig:teaser}
\end{figure}

Prior efforts to mitigate this fall into two camps. Post-hoc sampling strategies guide inference toward underrepresented modes~\cite{sehwag2022,um2024,um2025,morshed2025}; while effective for exploration, they leave the underlying distribution unchanged and do not prevent conditional mode collapse during training~\cite{aithal2024}. Training-time methods alter the loss or the conditioning pathway, e.g., aligning conditional and unconditional objectives so that head-class signal can transfer to the tail~\cite{ccua2025}, lesion-mask-controllable synthesis~\cite{zhang2025lefusion}, or causal disentanglement of pathology and anatomy~\cite{nie2025causal}. These approaches modify \emph{what} the model learns; they do not directly target the optimization-level pathology that gradient analyses on classification and language tasks identify as the underlying cause of long-tail collapse, namely that head- and tail-class gradients are mismatched in norm and direction and average destructively at the parameter level~\cite{francazi2023,dong2024}.
Dong et al.~\cite{dong2024} address this in long-tail language modeling by learning a cluster-conditional gating network that routes samples to specialized experts, but to our knowledge no analogous mechanism exists for long-tail text-to-image diffusion training, despite the fact that conditional flow matching is  the setting where conditioning-aligned routing is geometrically natural.

We bridge this gap with GRASP: Guided Residual Adapters with Sample-wise Partitioning. As displayed in Figure~\ref{fig:teaser}, GRASP uses a static external prior to partition training samples into groups in conditioning space (e.g., whole-label bins with a reserved healthy group, or clusters of text embeddings), so that each group's residual updates are isolated in its own expert.
During fine-tuning of a frozen pre-trained backbone, cluster-specific residual adapters (lightweight non-linear skips inserted into feedforward layers) enable targeted refinement without learned routing.
This preserves the original generation objective while enhancing stability and efficiency.

A natural objection is that this gives up the flexibility of learned gating. We argue the opposite holds in extreme long-tail.
A learned gate is itself dependent on a gradient signal dominated by head classes: with hundreds of head examples
per tail example, the gate has little incentive to route rare samples to dedicated experts.
A static, conditioning-driven partition removes this failure mode by construction: every tail sample is guaranteed to update its assigned expert, at the cost of fixing the partition from an external prior.

Our approach delivers clear and consistent improvements across challenging long-tail benchmarks.
On MIMIC-CXR-LT, GRASP reduces overall Fréchet Inception Distance (FID) by up to 80\% and increases tail-class diversity metrics such as coverage by up to 44\% compared to state-of-the-art baselines (Table~\ref{tab:cxr-metrics}).
When GRASP-generated images are used for downstream classification on NIH-CXR-LT, the macro-average F1 is comparable to real-data baseline while extending nonzero-F1 coverage from 3 of 13 classes (real data), and 1 of 13 (full fine-tuning) to 9 of 13 (Table~\ref{tab:cxr-f1-summary}).
Ablations indicate that both the partitioning and the residual adapter design contribute. The result does not depend on one grouping: random label-to-expert reassignments and other expert counts perform comparably, and only partitions that starve an expert degrade. With identical experts, fixed routing beats a learned gate on label-macro FID and downstream utility, while a gate initialized at the GRASP assignment finds no better alternative (Table~\ref{tab:partition-routing}). On the optimization trajectory, conditioning-based routing yields higher gradient consistency than a learned gate or a shared adapter (Figure~\ref{fig:grad_analysis}).

Our \textbf{contributions} are:
\textbf{(1)} We transfer the gradient-consistency objective of Dong et al.~\cite{dong2024}, originally proposed for long-tail language modeling with learned cluster-conditional gating, to long-tail text-to-image flow matching, and replace the learned gate with a deterministic, conditioning-driven partition. In conditional flow matching, condition values index distinct sets of probability paths, so partitioning along the conditioning is a structurally natural factorization rather than a generic embedding-space heuristic. We argue that, whenever even coarse priors over the conditioning are available or easily obtainable (which is the common case in medical imaging), a static partition that gives every expert sufficient support should be the default.
\textbf{(2)} GRASP is intentionally non-invasive: it leaves the flow-matching objective and the sampler unchanged, only restricting which parameters absorb which gradients. As a direct consequence it composes with orthogonal long-tail interventions: we demonstrate this by combining GRASP with self-guided minority sampling at inference time and observe that the composition favors static routing and outperforms minority guidance alone.
\textbf{(3)} We provide comprehensive validation across medical (MIMIC-CXR-LT and NIH-CXR-LT) and general (ImageNet-LT) benchmarks, including a gradient consistency analysis, a partitioning sensitivity study, and a capacity-matched comparison between fixed and learned routing, showing consistent gains in image quality and diversity over full fine-tuning, generic adapters, and learned-routing mixture-of-experts (MoE) variants under deliberately minimal priors.

\section{Related Work}
\label{sec:related_work}

\noindent\textbf{Long-tail diffusion training.}
Most prior work on long-tail text-to-image generation acts at inference time, steering sampling with classifier-free guidance (CFG) variants~\cite{dhariwal2021,rombach2022,ho2021} or minority-aware reweighting~\cite{sehwag2022,um2024,um2025,morshed2025,miao2024,zhang2024}. These methods can improve exploration but do not change the learned distribution and therefore cannot prevent training-time conditional collapse~\cite{aithal2024}. Training-time alternatives modify the target objective or conditioning pathway, including conditional-unconditional alignment (CCUA)~\cite{ccua2025}, failure-aware on-policy augmentation~\cite{park2025galtraj}, lesion-mask-controllable synthesis~\cite{zhang2025lefusion}, and causal pathology/anatomy disentanglement~\cite{nie2025causal}. GRASP is orthogonal: it keeps the flow-matching objective fixed and changes only which parameters absorb which gradients.

\noindent\textbf{Mixture-of-Experts and gradient consistency.}
Expert layers increase capacity through sparse routing~\cite{shazeer2017}, often with adapter-style experts such as LoRA~\cite{hu2022,wu2024mixture,li2025} or cluster-conditional gating as in MoCLE~\cite{gou2023}. Diffusion MoE variants similarly improve scaling or specialization through routed paths, timestep experts, dense-to-sparse conversion, or semantic superclass routing~\cite{xue2023raphael,zheng2025dense2moe,ganjdanesh2024diffpruning,wei2026promoe}. Our starting point is Cluster-Guided Sparse Experts~\cite{dong2024}, which optimizes intra-expert gradient consistency for long-tail language modeling with a learned gate. GRASP transfers this objective to flow-matching text-to-image diffusion, where conditions index probability paths, and replaces learned gating with a deterministic partition $\pi$. This guarantees that tail samples update their assigned expert even when head-class gradients dominate the training signal, a guarantee learned routers do not provide.

\noindent\textbf{Generative models for long-tail medical imaging.}
Medical diffusion work addresses rare pathology synthesis through text-guided augmentation~\cite{rajaraman2024addressing}, sparse feature-space mixing~\cite{elberg2024long}, lesion-focused objectives~\cite{zhang2025lefusion}, and causal disentanglement~\cite{nie2025causal}; adjacent systems scale chest-X-ray, segmentation, MRI, or 3D synthesis with larger domain-specific pipelines~\cite{chambon2024roentgen,zhang2025genseg,wu2025mrgen,guo2025maisi}. These approaches rely on stronger medical priors or large curated corpora. GRASP instead operates at fine-tuning time on a generic flow-matching backbone and is evaluated on both MIMIC-CXR-LT and ImageNet-LT, separating the routing mechanism from any single medical inductive bias.

\begin{figure*}[t]
    \centering
    \includegraphics[width=0.99\textwidth]{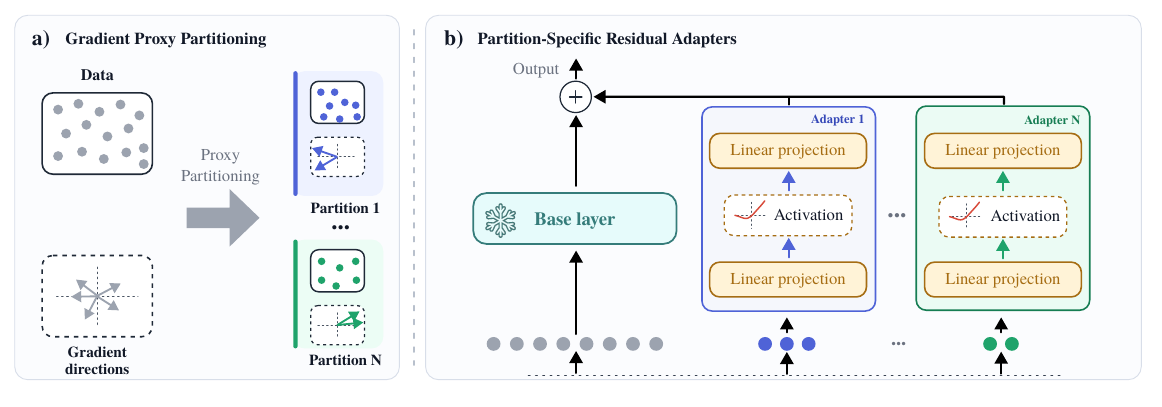}
    \caption{Overview of the GRASP architecture: \textbf{a)} We want to maximize gradient consistency during training by partitioning the samples so that each expert's dominant update direction aligns with that of the samples routed to it. We achieve this by partitioning based on flow-matching conditioning. \textbf{b)} Based on this partitioning, we deterministically route samples to their designated expert, while keeping the base model frozen.}
    \label{fig:method}
\end{figure*}
\section{Method}
\label{sec:method}

GRASP is a non-invasive architectural intervention. It leaves the flow-matching loss and the sampler unchanged and only modifies which parameters absorb which per-sample gradients. The contribution is therefore not a different generative objective but a routing strategy adapted to long-tail training.

\noindent\textbf{Sample-wise Partitioning.}
We measure how well an update serves a group of samples with the gradient consistency of~\citet{dong2024}(Figure~\ref{fig:method}):
\begin{equation}
\mathrm{GC}_{\theta}(X') =
\frac{g_{\theta}(X)\cdot g_{\theta}(X')}
{\lVert g_{\theta}(X)\rVert \,\lVert g_{\theta}(X')\rVert}
\label{eq:gc}
\end{equation}
where \(g_{\theta}(X')\) is the dominant update direction of a sample set $X'$, the leading singular vector of its per-sample gradients and $X$ is the set of samples that update the same parameters.
Directly optimizing this objective would require online computation and clustering of per-sample gradients, which is infeasible for large flow-matching transformers.

Instead of approximating gradient similarity directly, we partition the model's conditioning space. The key claim of this section is that, in conditional flow matching with class-style conditions, this is not a generic embedding-space heuristic: the conditioning is a structurally natural axis along which to factorize residual updates. This claim only concerns the domain of the factorization, so it does not assert that any particular partition is optimal, or that conditions grouped into one expert have similar gradients.

Conditional flow matching does not learn a single velocity field but a family of condition-indexed fields $\{v_\theta(\cdot,t,c)\}_{c\in\mathcal C}$, where each $c$ specifies a conditional target $p_{\mathrm{data}}(\cdot\mid c)$ and a probability path to the base distribution~\cite{lipman2022flow,albergo2023stochastic,kerrigan2024dcot,chemseddine2025conditionalwasserstein,atanackovic2025metaflow}. With class-style conditions, each value of $c$ thus indexes a \emph{distinct set of probability paths}. A static partition $\pi:\mathcal C \to \{1,\dots,K\}$ is the simplest mechanism that respects this structure: it groups samples by path family and gives each group a dedicated expert. The proxy claim is therefore not that semantically similar prompts have similar gradients in a generic embedding sense; it is that conditioning indexes path geometry in conditional flow matching, and we hypothesize that gradients inherit that geometry~\cite{cheng2025curse}. The partition need not be sharp: conditions in the same expert only need to be more path-homogeneous within than across, the regime in which the expert's residual adapter can absorb the shared local dynamics. Because consistency (Eq.~\ref{eq:gc}) compares each subset with the samples routed to the same expert, not with each other, an expert can also mix frequent and rare labels, as our label experts do to balance support (Table~\ref{tab:partition-composition}), and its rare labels can still score high. Figure~\ref{fig:grad_analysis} supports this on the optimization trajectory: GRASP reaches higher gradient consistency than MoCLE and a shared adapter, and expert-wise gradient consistency increases with expert-wise conditioning homogeneity.

Formally, let \(c_i \in \mathcal C\) denote the condition associated with sample \(x_i\). Depending on the setting, \(c_i\) may be obtained from a class-label embedding, \(c_i=\phi(\ell_i)\), or from a text/conditioning encoder, \(c_i=\psi(t_i)\). We define a static routing function
\begin{equation}
\pi:\mathcal C \rightarrow \{1,\dots,K\},
\end{equation}
which assigns each condition to one of \(K\) experts, and thus routes each sample \(x_i\) to expert \(\pi(c_i)\).

\begin{table}[!t]
\centering
\caption{Exact label-to-expert lookup $\pi$ used for MIMIC-CXR-LT (all 87,493 training samples). Ranks are pathology training-frequency ranks (1 = most frequent pathology, \textit{No Finding} excluded). Every label is assigned wholesale to one expert. E0--E3 correspond to C0--C3 in Figure~\ref{fig:clusters} and in Table~\ref{tab:expert_usage}.}
\label{tab:partition-composition}
\resizebox{\linewidth}{!}{%
\begin{tabular}{c r c p{0.62\linewidth}}
\toprule
\textbf{Expert} & \textbf{Samples} & \textbf{Ranks} & \textbf{Labels} \\
\midrule
E0 & 11,506 & 5--11 & Support Devices, Edema, Pneumonia, Pneumothorax, Lung Lesion, Fracture, Enlarged Cardiomediastinum \\
E1 & 9,243  & 1, 12--18 & Lung Opacity, Consolidation, Pleural Other, Calcification of the Aorta, Tortuous Aorta, Pneumoperitoneum, Subcutaneous Emphysema, Pneumomediastinum \\
E2 & 13,484 & 2--4 & Cardiomegaly, Atelectasis, Pleural Effusion \\
E3 & 53,260 & -- & No Finding (reserved healthy expert) \\
\bottomrule
\end{tabular}%
}
\end{table}

A straightforward instantiation is label-based partitioning which assigns every label wholesale to one expert, with several labels sharing one expert to ensure sufficient support. Medical imaging adds a domain-specific inductive bias: the healthy/pathological dichotomy. On MIMIC-CXR-LT~\cite{holste2022}, we therefore augment the partition with an additional cluster \(\pi_{\text{healthy}}\) containing healthy cases. Then we split the remaining 18 pathology labels into three groups, a number that follows the head/medium/tail convention~\cite{zhao2024,hou2025a}, giving $K = 4$ experts in total. To get non-starved experts, we use a balanced graph partition for construction, with the single labels treated as concept sets. This results in an approximately frequency-ordered whole label binning which serves as our $\pi$~(Table~\ref{tab:partition-composition}): ranked by training frequency, the first expert holds classes 2 to 4, the second 5 to 11 and the third 8 to 12, plus the largest pathology class (\emph{Lung Opacity}). The latter exception to the ordering reduces the imbalance factor from $16:1$ to $1.46:1$, giving 13,484/11,506/9,243 samples. This partition is merely used as a simple example, it was neither tuned nor curated. Section~\ref{sec:experiments} shows that random reassignments perform comparably.

While label-based routing is simple and interpretable, it ignores within-group conditioning variation. We therefore also consider text-based partitioning. Specifically, we map each sample to a semantic feature representation
$f(x_i)=\psi(t_i)$, where \(t_i\) is the paired textual prompt and \(\psi(\cdot)\) is the frozen text encoder of the diffusion backbone. We then compute \(\pi_{\text{text}}\) with bisecting \(K\)-means on \(\{f(x_i)\}\), using \(K=4\) and cosine similarity. In contrast to label-based routing, this groups samples by conditioning-space proximity rather than label identity or frequency alone. However, it is still a minimalist approach by design. The resulting cluster compositions are compared in Appendix~\ref{app:clusters}.

For ImageNet-LT, where label-level partitioning is impractical due to label cardinality and no analogue of the healthy/pathological dichotomy exists, we rely exclusively on text-embedding clustering with no domain-specific cluster. This is a deliberate choice: it shows that the routing framework generalizes beyond the medical setting and does not depend on handcrafted domain partitions, while still producing competitive long-tail behavior (Section~\ref{sec:experiments}).

We emphasize that both choices of $\pi$ are deliberately minimal. The label-based partition reuses the noisy labels released with MIMIC-CXR-LT, with a single domain-prior split (healthy vs.\ pathological) and a balance constraint; we do not denoise or curate them. The text-based partition is vanilla bisecting $K$-means on the frozen text-encoder embeddings, with no hand-tuned distance metric, weighting scheme, or post-hoc correction. The point of GRASP is not to engineer a high-quality prior. It is to show that even very coarse, off-the-shelf priors are sufficient to recover specialized adapters, as long as every expert receives sufficient support~(Section~\ref{sec:experiments}). This robustness positions the static partition as the default under extreme imbalance whenever such priors are at hand, which covers most medical imaging applications and is cheap to construct elsewhere via vanilla clustering; learned cluster-conditional gating remains the right choice when no usable prior exists. The deployment criterion is therefore practical: take whatever partition you can construct, give it to GRASP, and you receive the update isolation by construction, without modifying the loss or the sampler.

\noindent\textbf{Guided Residual Adapters.}
Given the static routing function \(\pi:\mathcal C \rightarrow \{1,\dots,K\}\), we condition the architecture via guided residual adapters. Unlike conventional MoE approaches with learned gating~\cite{shazeer2017,dong2024}, each sample \(x_i\) is deterministically assigned to expert \(\pi(c_i)\). This removes routing noise and makes specialization directly interpretable in terms of conditional subproblems.

The hidden representation at transformer block \(l\) is denoted by \(h_l \in \mathbb{R}^{d}\), and the corresponding base transformation by \(F_l(h_l)\). We augment this block with a residual adapter \(A_{k,l}\) specific to expert \(k=\pi(c_i)\). The layer output for sample \(x_i\) becomes
\begin{equation}
h_{l+1} = F_l(h_l) + A_{\pi(c_i),l}(h_l).
\label{eq:adapter_update}
\end{equation}
This yields a decomposition of the learned vector field into a shared backbone and an expert-specific residual correction,
\begin{equation}
\hat v_\theta(z,t,c)
=
s_\phi(z,t,c) + a_{\pi(c),\psi_{\pi(c)}}(z,t,c),
\label{eq:cond_grouped_decomposition}
\end{equation}
where \(s_\phi\) captures broadly shared dynamics and \(a_{\pi(c),\psi_{\pi(c)}}\) captures the correction associated with the selected condition group. The key assumption is not that all conditions within one expert are identical, but that after factoring out globally shared structure, the remaining residual dynamics are similar enough to justify parameter sharing within that expert.

Each adapter is a two-layer transformation $A_{k,l}(h) = W^{(2)}_{k,l}\,\sigma(W^{(1)}_{k,l} h)$ with expert-specific projections $(W^{(1)}_{k,l},W^{(2)}_{k,l})$. Residual adapters fit our setting because they combine strong parameter sharing with small condition-specific blocks~\cite{rebuffi2017,houlsby2019a,mahabadi2021hyperformerplusplus}: allocating a full expert per rare condition is statistically brittle, whereas grouped residual adapters share strength across related conditions while isolating condition-group-specific updates. In contrast to LoRA~\cite{hu2022}, which adds a linear low-rank update to existing weights, our adapter is a non-linear residual module whose width is not tied to a low rank constraint: $W^{(1)}_{k,l}\in\mathbb{R}^{d'\times d}$, $W^{(2)}_{k,l}\in\mathbb{R}^{d\times d'}$, with $d'$ ranging from $d'\ll d$ to $d'>d$. We empirically trade off adapter width $d'$ against the number of attached layers: either a few large adapters ($d' > d$) on the last dual-stream blocks (GRASP-HighDim) or lightweight adapters ($d' < d$) on all blocks (GRASP-LowDim).

By construction, expert-specific parameters $\psi_k$ are updated only by samples with $\pi(c)=k$, so $\nabla_{\psi_k}\mathcal L = \mathbb E[\mathbf 1\{\pi(C)=k\}\nabla_{\psi_k}\ell]$. This isolates group-specific gradients while the shared backbone $\phi$ provides transferable structure across conditions, the classical MoE intuition~\cite{jacobs1991adaptive}.

Our backbone is Flux~\cite{lipman2022flow,morshed2025}: a flow-matching transformer with 19 dual-stream blocks (text+image) followed by 38 single-stream blocks (image only), $L=57$ in total. Because dual blocks mediate multi-modal fusion, we instantiate $A_{k,l}$ in their FeedForward submodules; configurations that adapt all blocks (e.g., GRASP-LowDim) additionally attach $A_{k,l}$ to the MLP output projection of every single-stream block.

\section{Experiments}\label{sec:experiments}


\noindent\textbf{Datasets.} 
Motivated by our central goal of enabling more diverse and faithful image generation in medical imaging, we adopt MIMIC-CXR-LT~\cite{holste2022} as our primary benchmark. This dataset extends the large-scale MIMIC-CXR collection~\cite{johnson2019} by introducing additional rare disease categories and constructing single-label splits to emphasize long-tail behavior. The resulting distribution poses a realistic challenge for generative modeling, capturing the severe frequency imbalance observed in real clinical data.   
We follow the official MIMIC-CXR-LT protocol, using 87,493 training samples to train our text-to-image diffusion model and 23,550 test-set prompts to generate synthetic images for evaluation. Both splits exhibit similar long-tailed characteristics, enabling consistent assessment of performance across head and tail classes.

Likewise, we base the evaluation of the classifier in our downstream task on NIH-CXR-LT~\cite{holste2022}.
This chest X-ray dataset is a split of ChestXray14~\cite{wang2017} with additional long-tail classes added.
For evaluation, we take the proposed test split (20,279 images) and filter any labels that are not part of MIMIC-CXR-LT, resulting in 16,506 samples.
The downstream classifier is trained only on images generated from the MIMIC-CXR-LT test prompts and evaluated on this real NIH-CXR-LT set. We chose this cross-dataset design to limit memorization effects from report-to-image generation.

To show the generalizability of our method, we conduct further experiments on the widely used Imagenet-LT~\cite{liu2019} dataset. 
This allows us to verify our approach on a domain-unspecific long-tail setting. 
The train split consists of 115,846 image-label pairs, while the test split consists of 50,000 pairs.

\noindent\textbf{Metrics.}\label{subsec:metrics}
For the image generation, we want to reliably assess both the quality and diversity of the generated samples.
To quantify the quality of the synthetic images, we utilize the common metrics Fréchet Inception Distance (FID)~\cite{heusel2017}.
To assess the diversity of our generated samples, we employ Coverage~\cite{naeem2020}, which is generally more robust to outliers than Improved Recall~\cite{kynkaanniemi2019}. Given real features $\Phi_r$ and generated features $\Phi_g$, the k-nearest-neighbor radius around $x\in\Phi$ is $\rho_\Phi(x):=\|x-\mathrm{NN}_k(x;\Phi\setminus\{x\})\|$, and the corresponding ball-membership indicator is $f(x,\Phi):=\mathbf{1}[\exists y\in\Phi:\|x-y\|\le\rho_\Phi(y)]$. Coverage then estimates the fraction of real samples whose neighborhood contains a generated sample, $\mathrm{Cov}=N^{-1}\sum_{i=1}^N\max_j\mathbf{1}[\|\varphi_r^{(i)}-\varphi_g^{(j)}\|\le\rho_r(\varphi_r^{(i)})]$.
Since diversity is our primary concern, we report an additional robust diversity metric, called the adjusted Image Retrieval Score (IRS)~\cite{dombrowski2025}.
For each image, this intuitive metric tries to retrieve the closest image in the training and test datasets, estimating which percentage of the datasets the synthetic images cover.
The adjusted score is the test score normalized by the train score, which results in a metric that does not reward memorization, while keeping in mind the maximum possible coverage.

All image generation metrics are computed for a domain-specific DenseNet121~\cite{huang2017} as a feature extractor, which aligns with our downstream task and provides insights into radiographic properties of the generated images.
For broader comparability, we also report features computed by the well-studied feature extractor DINOv2~\cite{oquab2024}, giving us insights into domain-unspecific features of our synthetic dataset.
To gain robust estimates of our metrics, we generate 50,000 samples with resolution $1024 \times 1024$ per model and training seed.
For the evaluation of the classifier in the downstream task, we report the per-label results for the F1-score.

\noindent\textbf{Implementation.}
We fine-tune Flux~(FLUX.1-dev) with the flow-matching objective and classifier-free guidance (CFG) (scale~5) for 10,000 steps at batch size 8 on 8~H100 GPUs; full fine-tuning takes $\sim$30~h, adapters 7 to 12~h. The MoCLE baseline reuses our partitions and number of experts at LoRA rank 64. Per-seed downstream synthesis is 50,000 images at $1024^2$, the DenseNet-121 classifier is trained for 100 epochs over three classifier seeds and three dataset seeds (nine runs per configuration). Full setup and compute breakdown is in Appendix~\ref{app:compute}.

\noindent\textbf{Image Metrics Results.}
We evaluate generation quality and diversity in domain-specific (DenseNet, Table~\ref{tab:cxr-metrics}) and domain-unspecific (DINOv2, Table~\ref{tab:dino-metrics}) feature spaces. Adapter-based methods generally outperform full fine-tuning as well as CCUA, and within adapter-based methods static GRASP outperforms MoCLE and the shared adapter on aggregate metrics. The low-dimensional, all-layers GRASP variant gives the most uniform gains and the best global FID under CXR features; the high-dimensional, last-layer variant is slightly weaker globally but wins on some labels (e.g., \textit{Atelectasis}, \textit{Edema}). Under DINOv2 features the high-dimensional variant clearly leads.

The MoE-GRASP variants share GRASP's experts and capacity, but use a learned gate. They are behind unguided GRASP on label-macro FID in both feature spaces, and on most other metrics, while recovering less metrics downstream (Tables~\ref{tab:cxr-metrics},~\ref{tab:dino-metrics},~\ref{tab:cxr-f1-summary}). The learned variants only lead on two (LowDim) and four (HighDim) of the twelve feature-space metrics, mostly all-label ones. Additionally, their gates never make use of all experts and always route the healthy majority together with pathology groups. We read the loss in label-wise fidelity and tail utility as the signature of a gate that re-allocates routing capacity toward head-dominated gradients (Section~\ref{sec:intro}). CCUA~\cite{ccua2025}, which changes the training objective instead, performs worse than GRASP-LowDim on 11 of 12 feature space metrics and both downstream metrics. On ImageNet-LT, the high-dimensional GRASP variant again leads on aggregate; MoCLE recovers competitive performance on this dataset, indicating learned routing remains viable when no strong domain-prior partition is available.

\begin{table*}[!t]
\centering
\caption{CXR feature-space evaluation on MIMIC-CXR-LT. We report coverage (Cov., $\uparrow$), IRS ($\uparrow$), and FID ($\downarrow$) for all labels and macro averages. All-label $\pm$: SD over three generator seeds; macro $\pm$: SD across labels. Best values per column are shown in bold. \emph{Adapter} refers to fine-tuning using one shared sequential residual adapter without specific routing and Moe-GRASP refers to the learned matched comparison to our designs specified in Section~\ref{sec:method}.}
\label{tab:cxr-metrics}
\resizebox{0.92\textwidth}{!}{%
\begin{tabular}{
  c
  l
  *{2}{c c c}
}
\toprule
& & \multicolumn{3}{c}{\textbf{All labels}} & \multicolumn{3}{c}{\textbf{Macro Avg}} \\
\cmidrule(lr){3-5}\cmidrule(lr){6-8}
& \textbf{Method}
  & \textbf{Cov.} $\uparrow$ & \textbf{IRS} $\uparrow$ & \textbf{FID} $\downarrow$
  & \textbf{Cov.} $\uparrow$ & \textbf{IRS} $\uparrow$ & \textbf{FID} $\downarrow$ \\
\midrule
\multirow{7}{*}{\rotatebox[origin=c]{90}{\textbf{Baseline}}}
& \textbf{Full Fine-Tune (class)}              & $.55\!_{\pm\!.01}$  & $.11\!_{\pm\!.0}$   & $.06\!_{\pm\!.01}$  & $.62\!_{\pm\!.16}$  & $.17\!_{\pm\!.16}$  & $.14\!_{\pm\!.11}$ \\
& \textbf{Full Fine-Tune (text+class)}         & $.65\!_{\pm\!.01}$  & $.16\!_{\pm\!.01}$  & $.05\!_{\pm\!.01}$  & $.70\!_{\pm\!.13}$  & $.20\!_{\pm\!.15}$  & $.12\!_{\pm\!.10}$ \\
& \textbf{MoCLE}~\cite{gou2023}                & $.84\!_{\pm\!.01}$  & $.32\!_{\pm\!.01}$  & $.04\!_{\pm\!.0}$   & $.85\!_{\pm\!.12}$  & $.35\!_{\pm\!.17}$  & $.09\!_{\pm\!.07}$ \\
& \textbf{CCUA}~\cite{ccua2025}                & $.64\!_{\pm\!.03}$  & $.13\!_{\pm\!.01}$  & $.07\!_{\pm\!.0}$   & $.67\!_{\pm\!.15}$  & $.20\!_{\pm\!.18}$  & $.15\!_{\pm\!.11}$ \\
& \textbf{Adapter (768d)}                      & $.86\!_{\pm\!.01}$  & $.33\!_{\pm\!.02}$  & $.03\!_{\pm\!.0}$   & $.85\!_{\pm\!.12}$  & $.37\!_{\pm\!.17}$  & $.07\!_{\pm\!.07}$ \\
& \textbf{MoE-GRASP-LowDim}                    & $.927\!_{\pm\!.003}$ & $.454\!_{\pm\!.007}$ & $.012\!_{\pm\!.001}$ & $.903\!_{\pm\!.007}$ & $.498\!_{\pm\!.013}$ & $.051\!_{\pm\!.002}$ \\
& \textbf{MoE-GRASP-HighDim}                   & $.925\!_{\pm\!.004}$ & $.462\!_{\pm\!.010}$ & $.017\!_{\pm\!.002}$ & $.912\!_{\pm\!.003}$ & $.514\!_{\pm\!.015}$ & $.051\!_{\pm\!.001}$ \\
\midrule
\multirow{3}{*}{\rotatebox[origin=c]{90}{\textbf{Ours}}}
& \textbf{GRASP (768d)}                        & $.90\!_{\pm\!.01}$  & $.35\!_{\pm\!.01}$  & $.02\!_{\pm\!.0}$   & $\mathbf{.92\!_{\pm\!.10}}$  & $.53\!_{\pm\!.17}$  & $.05\!_{\pm\!.06}$ \\
& \textbf{GRASP-LowDim}                        & $.94\!_{\pm\!.0}$   & $.50\!_{\pm\!.0}$   & $\mathbf{.01\!_{\pm\!.0}}$   & $\mathbf{.92\!_{\pm\!.09}}$  & $.56\!_{\pm\!.12}$  & $\mathbf{.04\!_{\pm\!.05}}$ \\
& \textbf{GRASP-HighDim}                       & $.92\!_{\pm\!.01}$  & $.41\!_{\pm\!.01}$  & $.02\!_{\pm\!.0}$   & $.91\!_{\pm\!.10}$  & $.52\!_{\pm\!.10}$  & $\mathbf{.04\!_{\pm\!.05}}$ \\
\midrule
\multirow{3}{*}{\rotatebox[origin=c]{90}{\textbf{\,+\,MG}}}
& \textbf{Vanilla + MG}~\cite{um2024}                  & $.62\!_{\pm\!.02}$  & $.15\!_{\pm\!.01}$  & $.05\!_{\pm\!.01}$  & $.67\!_{\pm\!.02}$  & $.19\!_{\pm\!.02}$  & $.13\!_{\pm\!.01}$ \\
& \textbf{MoE-GRASP-LowDim + MG}~\cite{um2024}         & $.94\!_{\pm\!.0}$  & $.63\!_{\pm\!.02}$ & $\mathbf{.01\!_{\pm\!.0}}$  & $\mathbf{.92\!_{\pm\!.01}}$  & $.60\!_{\pm\!.01}$ & $.05\!_{\pm\!.0}$ \\
& \textbf{GRASP-LowDim + MG}~\cite{um2024}             & $\mathbf{.95\!_{\pm\!.0}}$  & $\mathbf{.68\!_{\pm\!.0}}$ & $\mathbf{.01\!_{\pm\!.0}}$  & $\mathbf{.92\!_{\pm\!.01}}$  & $\mathbf{.61\!_{\pm\!.0}}$ & $\mathbf{.04\!_{\pm\!.0}}$ \\
\bottomrule
\end{tabular}%
}
\end{table*}

\begin{table*}[!t]
\centering
\caption{DINOv2 feature-space evaluation on MIMIC-CXR-LT (all labels and macro over 19 labels) and ImageNet-LT (all labels); $\pm$ as in Table~\ref{tab:cxr-metrics}. Text-conditioned full fine-tuning does not apply to ImageNet-LT; CCUA, MoE-GRASP-HighDim and guided runs were evaluated on MIMIC-CXR-LT only. Best per column in bold.}
\resizebox{\textwidth}{!}{%
\begin{tabular}{
  c
  l
  *{3}{c c c}
}
\toprule
& & \multicolumn{6}{c}{\textbf{MIMIC-CXR-LT}} & \multicolumn{3}{c}{\textbf{ImageNet-LT}} \\
\cmidrule(lr){3-8}\cmidrule(lr){9-11}
& \textbf{Method}
  & \multicolumn{3}{c}{\textbf{All labels}}
  & \multicolumn{3}{c}{\textbf{Macro Avg}}
  & \multicolumn{3}{c}{\textbf{All labels}} \\
\cmidrule(lr){3-5}\cmidrule(lr){6-8}\cmidrule(lr){9-11}
& & \textbf{Cov.} $\uparrow$ & \textbf{IRS} $\uparrow$ & \textbf{FID} $\downarrow$
  & \textbf{Cov.} $\uparrow$ & \textbf{IRS} $\uparrow$ & \textbf{FID} $\downarrow$
  & \textbf{Cov.} $\uparrow$ & \textbf{IRS} $\uparrow$ & \textbf{FID} $\downarrow$ \\
\midrule
\multirow{7}{*}{\rotatebox[origin=c]{90}{\textbf{Baseline}}}
& \textbf{Full Fine-Tune (class)}              & $.01\!_{\pm\!.0}$   & $.08\!_{\pm\!.01}$ & $106.6\!_{\pm\!14.2}$ & $.07\!_{\pm\!.05}$ & $.22\!_{\pm\!.11}$ & $145.5\!_{\pm\!28.5}$ & $.71\!_{\pm\!.01}$  & $.76\!_{\pm\!.01}$ & $155.1\!_{\pm\!5.5}$ \\
& \textbf{Full Fine-Tune (text+class)}         & $.02\!_{\pm\!.01}$  & $.12\!_{\pm\!.02}$ & $88.3\!_{\pm\!12.6}$  & $.11\!_{\pm\!.07}$ & $.27\!_{\pm\!.14}$ & $128.3\!_{\pm\!29.1}$ & --                  & --                 & -- \\
& \textbf{MoCLE}~\cite{gou2023}                & $.02\!_{\pm\!.0}$   & $.13\!_{\pm\!.0}$  & $87.5\!_{\pm\!.43}$   & $.15\!_{\pm\!.11}$ & $.29\!_{\pm\!.14}$ & $123.0\!_{\pm\!29.2}$ & $\mathbf{.75\!_{\pm\!.0}}$   & $.76\!_{\pm\!.0}$  & $\mathbf{109.4\!_{\pm\!1.0}}$ \\
& \textbf{CCUA}~\cite{ccua2025}                & $.01\!_{\pm\!.0}$   & $.10\!_{\pm\!.01}$ & $117.7\!_{\pm\!39.5}$ & $.11\!_{\pm\!.08}$ & $.25\!_{\pm\!.10}$ & $157.5\!_{\pm\!26.4}$ & --  & --  & -- \\
& \textbf{Adapter (768d)}                      & $.02\!_{\pm\!.0}$   & $.15\!_{\pm\!.01}$ & $80.9\!_{\pm\!4.2}$   & $.14\!_{\pm\!.1}$  & $.34\!_{\pm\!.17}$ & $119.9\!_{\pm\!33.2}$ & $.57\!_{\pm\!.0}$   & $.39\!_{\pm\!.0}$  & $249.9\!_{\pm\!0.4}$ \\
& \textbf{MoE-GRASP-LowDim}                    & $.007\!_{\pm\!.000}$ & $.176\!_{\pm\!.002}$ & $98.6\!_{\pm\!3.8}$  & $.095\!_{\pm\!.010}$ & $.362\!_{\pm\!.017}$ & $155.2\!_{\pm\!7.3}$ & $.690\!_{\pm\!.004}$ & $\mathbf{.817\!_{\pm\!.002}}$ & $136.0\!_{\pm\!4.6}$ \\
& \textbf{MoE-GRASP-HighDim}                   & $.024\!_{\pm\!.003}$ & $.186\!_{\pm\!.002}$ & $73.4\!_{\pm\!2.1}$  & $.176\!_{\pm\!.017}$ & $.375\!_{\pm\!.005}$ & $114.5\!_{\pm\!1.7}$ & --  & --  & -- \\
\midrule
\multirow{3}{*}{\rotatebox[origin=c]{90}{\textbf{Ours}}}
& \textbf{GRASP (768d)}                        & $.02\!_{\pm\!.0}$   & $.17\!_{\pm\!.0}$  & $76.6\!_{\pm\!.54}$   & $.17\!_{\pm\!.12}$ & $.37\!_{\pm\!.17}$ & $115.4\!_{\pm\!39.2}$ & $.67\!_{\pm\!.07}$  & $.70\!_{\pm\!.22}$ & $164.4\!_{\pm\!60.2}$ \\
& \textbf{GRASP-LowDim}                        & $.01\!_{\pm\!.0}$   & $.17\!_{\pm\!.0}$  & $91.4\!_{\pm\!1.6}$   & $.12\!_{\pm\!.1}$  & $.38\!_{\pm\!.17}$ & $150.6\!_{\pm\!76.2}$ & $.70\!_{\pm\!.0}$   & $.81\!_{\pm\!.0}$  & $132.3\!_{\pm\!2.2}$ \\
& \textbf{GRASP-HighDim}                       & $\mathbf{.03\!_{\pm\!.0}}$   & $.19\!_{\pm\!.0}$  & $\mathbf{67.3\!_{\pm\!1.6}}$   & $\mathbf{.23\!_{\pm\!.16}}$ & $.39\!_{\pm\!.18}$ & $\mathbf{107.7\!_{\pm\!33.3}}$ & $\mathbf{.75\!_{\pm\!.0}}$   & $.80\!_{\pm\!.01}$ & $\mathbf{109.4\!_{\pm\!1.0}}$ \\
\midrule
\multirow{3}{*}{\rotatebox[origin=c]{90}{\textbf{\,+\,MG}}}
& \textbf{Vanilla + MG}~\cite{um2024}                  & $.01\!_{\pm\!.0}$   & $.11\!_{\pm\!.02}$ & $100.5\!_{\pm\!12.3}$ & $.08\!_{\pm\!.04}$ & $.25\!_{\pm\!.03}$ & $140.7\!_{\pm\!12.3}$ & --  & --  & -- \\
& \textbf{MoE-GRASP-LowDim + MG}~\cite{um2024}         & $.01\!_{\pm\!.0}$   & $\mathbf{.22\!_{\pm\!.0}}$ & $90.6\!_{\pm\!4.2}$   & $.11\!_{\pm\!.01}$ & $.41\!_{\pm\!.01}$ & $151.4\!_{\pm\!9.1}$ & --  & --  & -- \\
& \textbf{GRASP-LowDim + MG}~\cite{um2024}             & $.01\!_{\pm\!.0}$   & $.20\!_{\pm\!.0}$  & $84.0\!_{\pm\!1.6}$   & $.14\!_{\pm\!.0}$ & $\mathbf{.43\!_{\pm\!.0}}$ & $142.0\!_{\pm\!4.6}$ & --  & --  & -- \\
\bottomrule
\end{tabular}%
}
\label{tab:dino-metrics}
\end{table*}

\begin{figure*}[!t]
    \centering
    \includegraphics[width=\linewidth, clip]{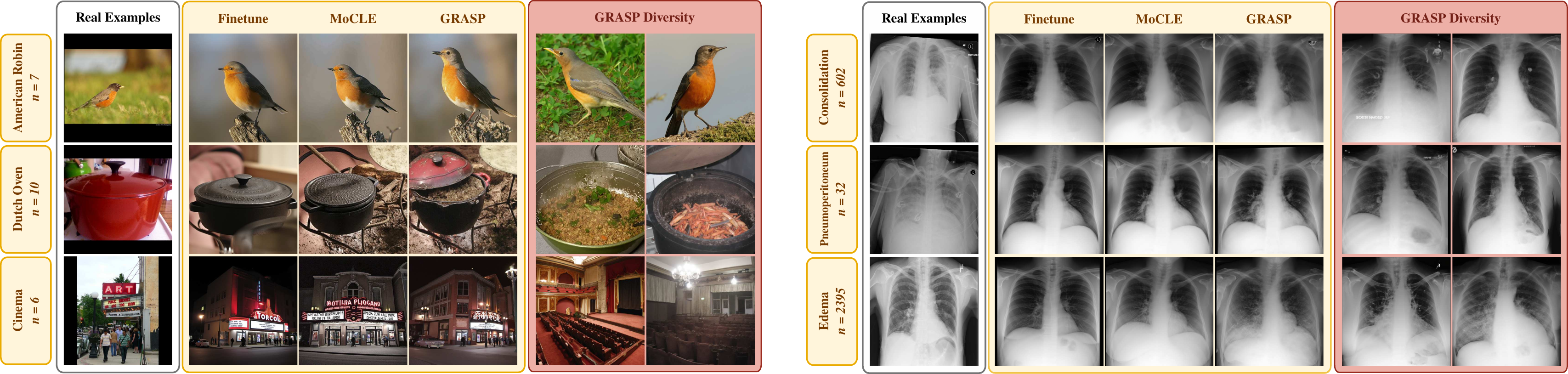}
    \caption{Qualitative comparison on tail classes of MIMIC-CXR-LT (random samples from heuristic pathologies) and ImageNet-LT (random pick from the bottom 2\%). Real counterparts are from NIH-CXR. Comparisons between vanilla, MoCLE~\cite{gou2023}, GRASP are made using the same seeds. The architectural intervention is intentionally non-invasive: GRASP outputs remain visually close to baselines on the same prompt, but show more distinct class-specific appearance that the baselines collapse on.}
    \label{fig:qualitative}
\end{figure*}

GRASP improves most metrics, especially on rare labels such as \textit{Pneumoperitoneum}; qualitatively, it preserves the baseline prompt structure while recovering class-specific features that baselines smooth out (Figure~\ref{fig:qualitative}).

\noindent\textbf{Composability with inference-time interventions.}
Because GRASP only changes which parameters take which gradients, it composes with self-guided minority sampling~\cite{um2024} (MG), a post-hoc sampler modification. With MG applied to both, fixed routing beats the learned router (MoE-GRASP-LowDim+MG) on 10 of 12 metrics, including every label-macro metric and CXR IRS (.68 vs. .63; Tables \ref{tab:cxr-metrics} and \ref{tab:dino-metrics}). GRASP-LowDim+MG attains the best all-labels IRS ($.68$) and macro IRS ($.61$) under CXR features (Table~\ref{tab:cxr-metrics}). MG on full fine-tune recovers only a small fraction of this gain, indicating that the lever is GRASP's training-time distribution shift, with MG re-allocating sampling effort within it. This is consistent with the intended role of static partitioning: it makes the architecture long-tail-friendly without precluding standard inference- or training-side recipes.

\begin{table*}[!t]
\centering
\caption{Partition sensitivity \textbf{(a)} and routing controls \textbf{(b)} on MIMIC-CXR-LT with fixed backbone, optimizer, schedule and evaluation budget. \textit{min.\ expert}: training samples of the smallest expert. The partition variants in (a) and the hybrid use one generator seed $\times$ three classifier seeds (GRASP and MoE-GRASP rows: three each). (b) compares learned with fixed routing at each adapter width; the fixed LowDim reference is the submitted partition in (a). Best value per column in bold. All 12 feature metrics: Appendix~\ref{app:partition}.}
\label{tab:partition-routing}
\resizebox{\textwidth}{!}{%
\begin{tabular}{l l c r cc cc c c}
\toprule
& & & & \multicolumn{2}{c}{\textbf{DINOv2 FID} $\downarrow$} & \multicolumn{2}{c}{\textbf{CXR FID} $\downarrow$} & & \\
\cmidrule(lr){5-6}\cmidrule(lr){7-8}
& \textbf{Configuration} & \textbf{$K$} & \textbf{min.\ expert} & \textbf{All} & \textbf{Macro} & \textbf{All} & \textbf{Macro} & \textbf{Macro F1} $\uparrow$ & \textbf{\#nz} $\uparrow$ \\
\midrule
\multirow{6}{*}{\rotatebox[origin=c]{90}{\textbf{(a) Partition}}}
& Submitted partition (GRASP-LowDim) & 4 & 9,243 & 91.4 & 150.6 & .014 & \textbf{.037} & .077 & \textbf{9} \\
& Random reassignment (mean of 3) & 4 & 7,927 & 84.2 & 139.6 & .012 & \textbf{.037} & .080 & \textbf{9} \\
& Fewer pathology groups & 3 & 17,080 & 89.5 & 149.7 & .014 & .041 & .075 & 8 \\
& More pathology groups & 5 & 8,431 & 94.2 & 150.7 & .014 & .039 & \textbf{.085} & 8 \\
& Literal official head/medium/tail~\cite{holste2022} & 3 & 71 & 80.6 & 317.6 & .018 & .075 & .071 & 2 \\
& One expert per label & 19 & 12 & 106.6 & 447.2 & \textbf{.011} & .050 & .073 & 5 \\
\midrule
\multirow{4}{*}{\rotatebox[origin=c]{90}{\textbf{(b) Routing}}}
& MoE-GRASP-LowDim (learned gate) & 4 & -- & 98.6 & 155.2 & .012 & .051 & .076 & 7 \\
& Static-initialized learned gate (LowDim) & 4 & -- & 94.8 & 150.0 & .013 & .038 & .081 & 6 \\
\cmidrule(lr){2-10}
& MoE-GRASP-HighDim (learned gate) & 4 & -- & 73.4 & 114.5 & .017 & .051 & .070 & 7 \\
& GRASP-HighDim (fixed) & 4 & 9,243 & \textbf{67.3} & \textbf{107.7} & .019 & .039 & .080 & 8 \\
\bottomrule
\end{tabular}%
}
\end{table*}

\noindent\textbf{Partition Sensitivity.}
Table~\ref{tab:partition-routing} swaps the submitted partition for alternatives of GRASP-LowDim width. Random reassignment randomizes only which labels share an expert. Its results are  comparable to the submitted version, so the results do not depend on our specific grouping. Using two or four pathology groups also stays close to ours, only versions that starve experts degrade: official head/medium/tail routing (a 71-sample tail expert) and one expert per label (12 samples for the rarest). Every configuration whose smallest expert holds sufficient support behaves like ours, so minimum expert support matters more than which labels share an expert. 

\noindent\textbf{Fixed Versus Learned Routing.} 
Table~\ref{tab:partition-routing} isolates routing: MoE-GRASP is identical to its corresponding GRASP variant, with the only difference being an added 512-parameter top-1 gate. In both widths, fixed routing improves label-macro FID in both feature spaces, label-macro DINOv2 coverage and IRS, all-label DINOv2 FID and the number of recovered classes, and matches or improves macro F1, while the learned gate is ahead mainly on all-label metrics; with
minority guidance applied to both, fixed routing also leads on CXR IRS. In a hybrid, we initialize the gate at
the GRASP assignment and train the adapters from scratch. The gate does not
drift across all checkpoints, and macro F1 is similar (.081), but recovery falls from 9/13 to 6/13. The learned gate
thus neither finds a better assignment nor improves on a viable static one.

\noindent\textbf{Downstream Classification Task.}
\label{subsec:classifier_results}
Table~\ref{tab:cxr-f1-summary} summarizes DenseNet-121 F1 on real NIH-CXR-LT images for classifiers trained only on synthetic images (nine runs per cell; the full per-label results in Appendix~\ref{app:classifier-table}). The task is intrinsically hard: even real MIMIC-CXR-LT training data yields nonzero F1 only on \textit{No Finding}, \textit{Pleural Effusion}, and \textit{Edema}. The informative result is therefore breadth rather than only macro F1. Full fine-tuning, CCUA and the shared adapter learn only the head class; MoCLE's learned cluster-conditional gate recovers limited tail signal. The static-partition GRASP variants improve further, producing nonzero mean F1 on eight or nine of thirteen classes while being comparable to the real-data macro F1. GRASP-LowDim recovers six classes on which real-data training has zero F1 and loses none in the reverse direction (one-sided: $p=.016$, two-sided $p=.031$). The remaining missed classes are extreme-tail labels with only a handful of training examples, where the real-data baseline also fails. Per-label gains over MoCLE are modest, often a few percentage points, and come with a small head-class reduction; the key result is that useful signal is spread across more labels. A paired two-sided Wilcoxon test on macro F1, paired by synthetic-dataset and classifier seed, separates GRASP-HighDim from every non-GRASP baseline ($p\!\leq\!0.031$), but not from the other static-partition GRASP variants ($p\!\geq\!0.20$).

\begin{table}[!t]
\centering
\small
\setlength{\tabcolsep}{3pt}

\begin{minipage}[c]{0.36\textwidth}
\captionof{table}{Downstream DenseNet-121 F1 on real NIH-CXR-LT (13 shared labels, 16,506 images), trained only on images generated from MIMIC-CXR-LT test prompts (Real: real MIMIC-CXR-LT images). \textbf{\#nz}: classes with nonzero mean F1. $p$: paired Wilcoxon over nine seed cells (exploratory). Learned-gate MoE-GRASP: Table~\ref{tab:partition-routing}(b). Per-label results: Appendix~\ref{app:classifier-table}.}
\label{tab:cxr-f1-summary}
\end{minipage}\hfill
\begin{minipage}[c]{0.61\textwidth}
\centering

\resizebox{\linewidth}{!}{%
\begin{tabular}{l c c c c}
\toprule
\textbf{Training data} & \textbf{\#nz} & \textbf{Macro F1 (13)} $\uparrow$ & \textbf{No Finding} & \textbf{$p$ vs GRASP-HD} \\
\midrule
Real MIMIC-CXR-LT (reference) & 3 & $.079\!_{\pm\!.007}$ & $.881\!_{\pm\!.002}$ & -- \\
\midrule
FT (class)             & 1 & $.068\!_{\pm\!.000}$ & $\mathbf{.882\!_{\pm\!.000}}$ & $.004$ \\
FT (text+class)        & 1 & $.068\!_{\pm\!.000}$ & $\mathbf{.882\!_{\pm\!.000}}$ & $.004$ \\
MoCLE~\cite{gou2023}   & 5 & $.069\!_{\pm\!.002}$ & $.881\!_{\pm\!.001}$ & $.004$ \\
CCUA~\cite{ccua2025}   & 1 & $.068\!_{\pm\!.000}$ & $\mathbf{.882\!_{\pm\!.000}}$ & $.004$ \\
Adapter (768d)         & 1 & $.068\!_{\pm\!.000}$ & $\mathbf{.882\!_{\pm\!.000}}$ & $.031$ \\
\midrule
GRASP (768d)           & 9 & $.076\!_{\pm\!.004}$ & $.872\!_{\pm\!.006}$ & $.203$ \\
GRASP-LowDim           & 9 & $.077\!_{\pm\!.005}$ & $.868\!_{\pm\!.010}$ & $.359$ \\
GRASP-HighDim          & 8 & $\mathbf{.080\!_{\pm\!.006}}$ & $.848\!_{\pm\!.020}$ & -- \\
\bottomrule
\end{tabular}%
}
\end{minipage}

\end{table}

\begin{table}[!t]
\centering
\caption{Ablation on resampling and clustering. Higher is better for Cov/IRS; lower is better for FID.}
\label{tab:resample-ablation}
\resizebox{0.5\columnwidth}{!}{%
\begin{tabular}{c c c c c c}
\toprule
\textbf{Label} & \textbf{Resample} & \textbf{Embed} & \textbf{Cov} $\uparrow$ & \textbf{IRS} $\uparrow$ & \textbf{FID} $\downarrow$ \\
\midrule
$\checkmark$ & $-$ & $-$ & $.91\!_{\pm\!.1}$ & $.52\!_{\pm\!.1}$ & $\mathbf{.04\!_{\pm\!.05}}$ \\
 $\checkmark$ & $\checkmark$ & $-$ & $\mathbf{.92\!_{\pm\!.09}}$ & $\mathbf{.52\!_{\pm\!.12}}$ & $\mathbf{.04\!_{\pm\!.05}}$ \\
$-$  & $\checkmark$ & $\checkmark$ & $.91\!_{\pm\!.1}$ & $.52\!_{\pm\!.16}$ & $.05\!_{\pm\!.06}$ \\
\bottomrule
\end{tabular}%
}
\end{table}

\noindent\textbf{Partitioning Source and Resampling.}
\label{subsec:part_n_res}
Table~\ref{tab:resample-ablation} ablates partition choice and round-robin resampling, which guarantees one sample per expert per step. Label partitions slightly outperform vanilla $K$-means, and resampling gives a small diversity boost at no quality cost; we keep it off in headline runs to keep the method minimal but recommend enabling it.

\noindent\textbf{Gradient Analysis.}
Figure~\ref{fig:grad_analysis} supports the mechanism on the optimization trajectory: GRASP has the highest within-expert gradient consistency over the first 1,000 steps, with MoCLE intermediate and the single-adapter baseline lowest. Checkpoint-wise conditioning homogeneity also correlates positively with expert-wise gradient alignment.

\input{figures/grad_analysis}

\noindent\textbf{Limitations.}
Downstream gains remain modest because the classifier task itself is extremely long-tailed, as shown by the real-data baseline collapsing on most labels. Also, the partition $\pi$ is fixed before training and cannot be corrected from gradients; we accept this for the tail-routing guarantee, and empirically GRASP tolerates coarse partitions, including random label reassignments and vanilla $K$-means embeddings (Tables~\ref{tab:partition-routing} and~\ref{tab:resample-ablation}). Finally, the FLUX-scale backbone makes the fine-tuning compute-intensive (Appendix~\ref{app:compute}, whereas GRASP itself adds little overhead.

\section{Conclusion}
\label{sec:conclusion}
GRASP is a non-invasive intervention for long-tail text-to-image flow matching. A static conditioning-space partition routes samples to group-specific residual adapters, improving gradient consistency without changing the objective or sampler. Unlike a learned gate, this routing guarantee does not depend on an optimization signal dominated by head classes, which is the failure mode we target in extreme long-tail regimes. Across MIMIC-CXR-LT, NIH-CXR-LT, and ImageNet-LT, GRASP improves rare-class quality and diversity, supports stronger downstream classifiers than non-GRASP synthetic baselines, and composes with inference-time minority guidance. Learned routing remains appropriate when no usable prior exists, but when such priors are available, static routing is the more stable default.

\section*{Funding}
The authors gratefully acknowledge the scientific support and HPC resources provided by the Erlangen National High Performance Computing Center (NHR@FAU) of the Friedrich-Alexander-Universit\"at Erlangen-N\"urnberg (FAU) under the NHR project b180dc. NHR funding is provided by federal and Bavarian state authorities. NHR@FAU hardware is partially funded by the German Research Foundation (DFG) -- 440719683.

\section*{Competing Interests} The authors declare no competing interests.

\section*{Ethics Statement}
This work uses only de-identified, publicly available research datasets. MIMIC-CXR and its long-tailed split MIMIC-CXR-LT~\cite{johnson2019,holste2022} were accessed through PhysioNet under its credentialed-access data use agreement; NIH ChestX-ray14 / NIH-CXR-LT~\cite{wang2017,holste2022} and ImageNet-LT~\cite{liu2019} are publicly released. No new data were collected from human subjects. 
Improving rare-condition synthesis could make model development and stress testing less dependent on scarce medical data and may help expose failures on underrepresented pathologies. However, synthetic images are not a substitute for clinically grounded cohorts: they can preserve or amplify dataset biases, encode spurious correlations, and create misleading evidence if used without provenance and downstream validation. Our evaluation relies on feature-space metrics and a downstream classifier, not on expert radiological reading, so it does not establish clinical fidelity of individual images. We therefore frame GRASP as a controlled augmentation and analysis tool, not as an automatic replacement for real data or expert review, and synthetic images produced with it should not be used for clinical decision-making.

\bibliographystyle{plainnat}
\bibliography{main}

\appendix

\section{Cluster compositions}
\label{app:clusters}

We visualize the cluster compositions induced by the two partition strategies in Figure~\ref{fig:clusters}: the label-based partition (frequency-binned MIMIC-CXR labels with a healthy split) and the text-embedding partition (vanilla bisecting $K$-means on frozen text-encoder features). The label-based partition assigns each sample to a single cluster by frequency band, while the text-based partition produces softer, conditioning-space groupings whose label distributions overlap.

\begin{figure}[htbp]
    \centering

\definecolor{pNoFind}{HTML}{1F77B4}
\definecolor{pLungOpac}{HTML}{AEC7E8}
\definecolor{pCardio}{HTML}{FF7F0E}
\definecolor{pAtelect}{HTML}{FFBB78}
\definecolor{pPleEff}{HTML}{2CA02C}
\definecolor{pSupDev}{HTML}{98DF8A}
\definecolor{pEdema}{HTML}{D62728}
\definecolor{pPneum}{HTML}{FF9896}
\definecolor{pOther}{HTML}{9EDAE5}

\newcommand{\ClusterPie}[5]{%
  \begin{scope}[shift={(#1,#2)}]
    \pie[
      radius=#3,
      hide number,
      hide label,
      style=draw=none,
      color={pNoFind,pLungOpac,pCardio,pAtelect,pPleEff,pSupDev,pEdema,pPneum,pOther},
      sum=100
    ]{#5}
    \pgfmathsetmacro{\LabelY}{-(#3 + 0.30)}
    \node[font=\tiny, align=center] at (0,\LabelY) {#4};
  \end{scope}%
}

\newcommand{\ClusterSolid}[4]{%
  \begin{scope}[shift={(#1,#2)}]
    \fill[pNoFind] (0,0) circle (#3);
    \draw[black, thin] (0,0) circle (#3);
    \pgfmathsetmacro{\LabelY}{-(#3 + 0.30)}
    \node[font=\tiny, align=center] at (0,\LabelY) {#4};
  \end{scope}%
}

\resizebox{0.7\linewidth}{!}{%
\begin{tikzpicture}[x=1cm,y=1cm]

  \node[font=\small, anchor=west] at (0.2, 2.1) {\textbf{Label Partitions}};
  
  \ClusterPie{0.8}{1.2}{0.31}{C0\\(11,506)}{
    0/NoF,0/LO,0/Card,0/Ate,0/Ple,28.49817/Sup,20.81523/Ede,19.07700/Pne,31.60959/Oth
  }
  \ClusterPie{1.8}{1.2}{0.29}{C1\\(9,243)}{
    0/NoF,85.76220/LO,0/Card,0/Ate,0/Ple,0/Sup,0/Ede,0/Pne,14.23780/Oth
  }
  \ClusterPie{3.0}{1.2}{0.34}{C2\\(13,484)}{
    0/NoF,0/LO,37.91902/Card,33.66212/Ate,28.41887/Ple,0/Sup,0/Ede,0/Pne,0/Oth
  }
  \ClusterSolid{4.8}{1.2}{0.68}{C3\\(53,260)}

  \node[font=\small, anchor=west] at (0.2, -0.2) {\textbf{Text Cluster Partitions}};

  \ClusterPie{0.8}{-0.9}{0.40}{C0\\(18,256)}{
    29.80938/NoF,11.68383/LO,8.39724/Card,13.74890/Ate,11.91389/Ple,1.02980/Sup,8.85188/Ede,6.41981/Pne,8.14527/Oth
  }
  \ClusterPie{2.0}{-0.9}{0.47}{C1\\(25,454)}{
    24.33802/NoF,21.50939/LO,10.57594/Card,7.49980/Ate,6.39585/Ple,11.07095/Sup,2.47505/Ede,3.55936/Pne,12.57563/Oth
  }
  \ClusterPie{3.3}{-0.9}{0.47}{C2\\(25,846)}{
    97.64374/NoF,0.03869/LO,1.79138/Card,0.10833/Ate,0.04256/Ple,0/Sup,0.03095/Ede,0.13542/Pne,0.20893/Oth
  }
  \ClusterPie{4.6}{-0.9}{0.39}{C3\\(17,937)}{
    91.35307/NoF,1.72270/LO,2.36940/Card,0.51291/Ate,0.10035/Ple,1.52199/Sup,0.78608/Ede,0.45716/Pne,1.17634/Oth
  }

  \begin{scope}[shift={(6.5, 1.4)}]
    \node[font=\scriptsize, anchor=west] at (0, 0.4) {\textbf{Findings}};
    \fill[pNoFind] (0.2, 0.0) rectangle +(0.25,0.25); \node[font=\tiny, anchor=west] at (0.6, 0.12) {No Finding};
    \fill[pLungOpac] (0.2,-0.4) rectangle +(0.25,0.25); \node[font=\tiny, anchor=west] at (0.6,-0.28) {Lung Opacity};
    \fill[pCardio] (0.2,-0.8) rectangle +(0.25,0.25); \node[font=\tiny, anchor=west] at (0.6,-0.68) {Cardiomegaly};
    \fill[pAtelect] (0.2,-1.2) rectangle +(0.25,0.25); \node[font=\tiny, anchor=west] at (0.6,-1.08) {Atelectasis};
    \fill[pPleEff] (0.2,-1.6) rectangle +(0.25,0.25); \node[font=\tiny, anchor=west] at (0.6,-1.48) {Pleural Effusion};
    \fill[pSupDev] (0.2,-2.0) rectangle +(0.25,0.25); \node[font=\tiny, anchor=west] at (0.6,-1.88) {Support Devices};
    \fill[pEdema] (0.2,-2.4) rectangle +(0.25,0.25); \node[font=\tiny, anchor=west] at (0.6,-2.28) {Edema};
    \fill[pPneum] (0.2,-2.8) rectangle +(0.25,0.25); \node[font=\tiny, anchor=west] at (0.6,-2.68) {Pneumonia};
    \fill[pOther] (0.2,-3.2) rectangle +(0.25,0.25); \node[font=\tiny, anchor=west] at (0.6,-3.08) {Other};
  \end{scope}

\end{tikzpicture}
}
    \caption{Composition of the partitioning based on labels (top) compared to the partitioning based on text clusters (bottom).}
    \label{fig:clusters}
\end{figure}

\section{Expert specialization and resampling}
\label{app:expert-spec}

Figure~\ref{fig:irs_spec_delta}(a) reports per-expert label-wise diversity for the label-based partition. Experts inherit the conditioning composition of their cluster: the expert assigned to the (single-label) healthy cluster performs best on \textit{No Finding} and worse on most pathologies, while mixed experts perform best on the labels that dominate their cluster (e.g., the Expert~0 cluster has the largest share of \textit{Enlarged Mediastinum} and \textit{Fracture}, and Expert~0 is best on those labels). The relationship is consistent but noisy, with occasional outliers. Figure~\ref{fig:irs_spec_delta}(b) shows the effect of resampling on expert-wise diversity: resampling redistributes utilization (Table~\ref{tab:expert_usage}) and yields a small but consistent overall improvement, at the cost of a slight per-expert quality reduction in the previously dominant healthy expert.


\begin{figure}[!t]
  \centering

  \begin{subfigure}[t]{0.70\linewidth}
    \centering
    \begin{tikzpicture}
      \begin{axis}[
        width=\linewidth,
        height=0.55\linewidth,
        ybar,
        bar width=1.9pt,
        ymin=0,
        ymax=1.0,
        xmin=-0.8,
        xmax=18.8,
        ylabel={IRS},
        ylabel style={font=\tiny, scale=0.9},
        xtick={0, 1, 2, 3, 4, 5, 6, 7, 8, 9, 10, 11, 12, 13, 14, 15, 16, 17, 18},
        xticklabels={Atelect., Calc. Aorta, Cardiomeg., Consol., Edema, Enlarged Cmd., Fracture, Lung Les., Lung Opac., No Finding, Pl. Effus., Pl. Other, Pneumomed., Pneumonia, Pneumoperit., Pneumothor., Subcut. Emph., Sup. Devices, Tort. Aorta},
        x tick label style={rotate=60, anchor=east, font=\tiny, scale=0.82},
        y tick label style={font=\tiny, scale=0.9},
        grid=major,
        grid style={dashed,gray!30},
        legend style={draw=none, font=\tiny, scale=0.85, inner sep=0.5pt, row sep=0pt, at={(0.5,0.98)}, anchor=north, fill=white, fill opacity=0.6, text opacity=1},
        legend columns=4,
        /tikz/every even column/.append style={column sep=1pt},
        legend image code/.code={\draw[#1,draw=none] (0cm,-0.08cm) rectangle (0.18cm,0.08cm);},
        tick align=outside,
        enlarge x limits=0.01
      ]
        \addplot+[fill=blue!65!black, draw=none, bar shift=-2.85pt] coordinates {(0,0.433000) (1,0.578000) (2,0.384000) (3,0.606000) (4,0.658000) (5,0.598000) (6,0.393000) (7,0.505000) (8,0.504000) (9,0.236000) (10,0.268000) (11,0.343000) (12,0.500000) (13,0.478000) (14,0.630000) (15,0.342000) (16,0.520000) (17,0.578000) (18,0.631000)};
        \addplot+[fill=orange!90!black, draw=none, bar shift=-0.95pt] coordinates {(0,0.446000) (1,0.439000) (2,0.345000) (3,0.512000) (4,0.523000) (5,0.548000) (6,0.442000) (7,0.458000) (8,0.576000) (9,0.242000) (10,0.266000) (11,0.366000) (12,0.500000) (13,0.505000) (14,0.741000) (15,0.432000) (16,0.520000) (17,0.570000) (18,0.598000)};
        \addplot+[fill=green!60!black, draw=none, bar shift=0.95pt] coordinates {(0,0.551000) (1,0.613000) (2,0.429000) (3,0.430000) (4,0.601000) (5,0.705000) (6,0.520000) (7,0.547000) (8,0.557000) (9,0.293000) (10,0.321000) (11,0.323000) (12,0.500000) (13,0.551000) (14,0.630000) (15,0.376000) (16,0.560000) (17,0.619000) (18,0.656000)};
        \addplot+[fill=red!70!black, draw=none, bar shift=2.85pt] coordinates {(0,0.138000) (1,0.150000) (2,0.150000) (3,0.147000) (4,0.131000) (5,0.345000) (6,0.271000) (7,0.219000) (8,0.164000) (9,0.304000) (10,0.100000) (11,0.110000) (12,0.917000) (13,0.219000) (14,0.444000) (15,0.113000) (16,0.240000) (17,0.273000) (18,0.295000)};
        \legend{E0,E1,E2,E3}
      \end{axis}
    \end{tikzpicture}
    \caption{Expert specialization shown by label-wise IRS scores for each expert (no-resample setting).}
  \end{subfigure}\hfill
  \begin{subfigure}[t]{0.26\linewidth}
    \centering
    \pgfplotstableread[row sep=\\, col sep=space]{%
      x y delta \\
0 0 -0.024000 \\
1 0 -0.089000 \\
2 0 -0.015000 \\
3 0 -0.046000 \\
0 1 -0.075000 \\
1 1 -0.174000 \\
2 1 0.070000 \\
3 1 -0.070000 \\
0 2 -0.021000 \\
1 2 -0.030000 \\
2 2 -0.063000 \\
3 2 -0.044000 \\
0 3 0.058000 \\
1 3 0.024000 \\
2 3 -0.106000 \\
3 3 -0.018000 \\
0 4 0.070000 \\
1 4 -0.018000 \\
2 4 -0.047000 \\
3 4 -0.026000 \\
0 5 -0.269000 \\
1 5 -0.192000 \\
2 5 0.050000 \\
3 5 -0.141000 \\
0 6 -0.148000 \\
1 6 0.007000 \\
2 6 0.063000 \\
3 6 -0.149000 \\
0 7 0.029000 \\
1 7 0.060000 \\
2 7 0.062000 \\
3 7 -0.102000 \\
0 8 -0.016000 \\
1 8 0.043000 \\
2 8 0.041000 \\
3 8 -0.036000 \\
0 9 -0.035000 \\
1 9 -0.046000 \\
2 9 -0.008000 \\
3 9 -0.080000 \\
0 10 -0.028000 \\
1 10 -0.012000 \\
2 10 -0.008000 \\
3 10 -0.017000 \\
0 11 -0.074000 \\
1 11 0.000000 \\
2 11 0.051000 \\
3 11 -0.051000 \\
0 12 -0.083000 \\
1 12 -0.083000 \\
2 12 -0.083000 \\
3 12 -0.083000 \\
0 13 -0.095000 \\
1 13 -0.046000 \\
2 13 -0.035000 \\
3 13 -0.075000 \\
0 14 0.074000 \\
1 14 -0.074000 \\
2 14 0.074000 \\
3 14 0.000000 \\
0 15 -0.004000 \\
1 15 0.121000 \\
2 15 0.034000 \\
3 15 -0.032000 \\
0 16 0.000000 \\
1 16 0.040000 \\
2 16 0.000000 \\
3 16 0.000000 \\
0 17 -0.060000 \\
1 17 -0.008000 \\
2 17 -0.029000 \\
3 17 -0.085000 \\
0 18 -0.230000 \\
1 18 -0.148000 \\
2 18 -0.164000 \\
3 18 -0.262000 \\
    }\deltatable
    \begin{tikzpicture}
      \begin{axis}[
        width=0.36\linewidth,
        height=1.48\linewidth,
        scale only axis,
        xlabel={},
        ylabel={},
        xmin=-0.5,
        xmax=3.5,
        ymin=-0.5,
        ymax=18.5,
        y dir=reverse,
        xtick={0,1,2,3},
        xticklabels={E0,E1,E2,E3},
        x tick label style={font=\tiny, scale=0.82},
        ytick={0, 1, 2, 3, 4, 5, 6, 7, 8, 9, 10, 11, 12, 13, 14, 15, 16, 17, 18},
        yticklabels={Atelect., Calc. Aorta, Cardiomeg., Consol., Edema, Enlarged Cmd., Fracture, Lung Les., Lung Opac., No Finding, Pl. Effus., Pl. Other, Pneumomed., Pneumonia, Pneumoperit., Pneumothor., Subcut. Emph., Sup. Devices, Tort. Aorta},
        y tick label style={font=\tiny, scale=0.82},
        tick align=outside,
        axis on top,
        colormap={rwb}{rgb255=(33,102,172) rgb255=(247,247,247) rgb255=(178,24,43)},
        point meta min=-0.269000,
        point meta max=0.269000,
        colorbar,
        colorbar style={y tick label style={font=\tiny, scale=0.82}, width=1.8mm},
      ]
        \addplot[
          matrix plot*,
          mesh/cols=4,
          point meta=explicit,
          draw=none
        ] table[x=x,y=y,meta=delta] {\deltatable};
      \end{axis}
    \end{tikzpicture}
    \caption{$\Delta$IRS (no-resample $-$ resample).}
  \end{subfigure}

  \caption{Comparison of expert specialization and resampling impact.}
  \label{fig:irs_spec_delta}
\end{figure}

\begin{table}[htbp]
\centering
\caption{Expert usage percentages by partitioning and sampling method for MIMIC-CXR-LT.}
\label{tab:expert_usage}
\resizebox{\linewidth}{!}{%
\begin{tabular}{@{}ccc*{4}{S[table-format=2.1]}}
\toprule
\multicolumn{1}{c}{\textbf{Label}} & \multicolumn{1}{c}{\textbf{Embed}} & \multicolumn{1}{c}{\textbf{Resample}} & \multicolumn{1}{c}{\textbf{Expert 0 (\%)}} & \multicolumn{1}{c}{\textbf{Expert 1 (\%)}} & \multicolumn{1}{c}{\textbf{Expert 2 (\%)}} & \multicolumn{1}{c}{\textbf{Expert 3 (\%)}} \\
\midrule
\checkmark &  &  & 13.01 & 10.34 & 15.44 & 61.20 \\
\checkmark & & \checkmark & 24.03 & 19.78 & 17.08 & 39.11 \\
& \checkmark &  & 21.82 & 28.79 & 29.29 & 20.10 \\
& \checkmark & \checkmark & 25.64 & 26.46 & 25.81 & 22.09 \\
\bottomrule
\end{tabular} %
}
\end{table}

\section{Per-label downstream classification results}
\label{app:classifier-table}

Table~\ref{tab:cxr-f1} reports the full per-label F1 scores summarized in Table~\ref{tab:cxr-f1-summary} of the main text. We include this table for completeness; the main-text discussion (Section~\ref{subsec:classifier_results}) is based on these values, in particular the nonzero-coverage count (\textbf{\#nz}) and the macro F1 over all 13 classes. The four classes \textit{Calc.\ of the Aorta}, \textit{Pneumomediastinum}, \textit{Pneumoperitoneum}, and \textit{Tortuous Aorta} collapse to true zero F1 across every method (including the real-data baseline), reflecting the extreme tail of the MIMIC-CXR-LT distribution.

\begin{table*}[!t]
\centering
\caption{Downstream DenseNet-121 classification, full per-label F1 on the real NIH-CXR-LT test set restricted to the 13 labels shared with MIMIC-CXR-LT (16,506 images). Classifiers are trained only on images generated from the 23,550 MIMIC-CXR-LT test prompts (Real: real MIMIC-CXR-LT images). Mean$\pm$SD across the nine runs per cell (three synthetic-dataset seeds $\times$ three classifier seeds; three classifier seeds for Real). Best values per row are shown in bold. Entries reported as $.000\!_{\pm\!.000}$ are true zeros (no correct predictions across any of the nine runs), not rounded near-zeros. The body summary view is in Table~\ref{tab:cxr-f1-summary}.}
\label{tab:cxr-f1}
\resizebox{\textwidth}{!}{%
\begin{tabular}{
  l
  *{11}{c}
}
\toprule
& \multicolumn{1}{c}{\textbf{Real}} & \multicolumn{5}{c}{\textbf{Baselines}} & \multicolumn{2}{c}{\textbf{Learned gate}} & \multicolumn{3}{c}{\textbf{GRASP}} \\
\cmidrule(lr){2-2}\cmidrule(lr){3-7}\cmidrule(lr){8-9}\cmidrule(lr){10-12}
\textbf{Label} &
\textbf{Real} &
\textbf{FT (class)} &
\textbf{FT (text+class)} &
\textbf{MoCLE} &
\textbf{CCUA} &
\textbf{Adapter (768d)} &
\textbf{MoE-GRASP-LowDim} &
\textbf{MoE-GRASP-HighDim} &
\textbf{GRASP (768d)} &
\textbf{GRASP-LowDim} &
\textbf{GRASP-HighDim} \\
\midrule
No Finding & $.881\!_{\pm\!.002}$ & $\mathbf{.882\!_{\pm\!.000}}$ & $\mathbf{.882\!_{\pm\!.000}}$ & $.881\!_{\pm\!.001}$ & $\mathbf{.882\!_{\pm\!.000}}$ & $\mathbf{.882\!_{\pm\!.000}}$ & $.870\!_{\pm\!.005}$ & $.879\!_{\pm\!.004}$ & $.872\!_{\pm\!.006}$ & $.868\!_{\pm\!.010}$ & $.848\!_{\pm\!.020}$ \\
Pleural Effusion & $\mathbf{.130\!_{\pm\!.091}}$ & $.000\!_{\pm\!.000}$ & $.000\!_{\pm\!.000}$ & $.004\!_{\pm\!.004}$ & $.000\!_{\pm\!.000}$ & $.000\!_{\pm\!.000}$ & $.081\!_{\pm\!.046}$ & $.013\!_{\pm\!.019}$ & $.023\!_{\pm\!.029}$ & $.052\!_{\pm\!.046}$ & $.082\!_{\pm\!.061}$ \\
Edema & $.011\!_{\pm\!.009}$ & $.000\!_{\pm\!.000}$ & $.000\!_{\pm\!.000}$ & $.012\!_{\pm\!.023}$ & $.000\!_{\pm\!.000}$ & $.000\!_{\pm\!.000}$ & $.008\!_{\pm\!.014}$ & $.005\!_{\pm\!.014}$ & $\mathbf{.061\!_{\pm\!.027}}$ & $.032\!_{\pm\!.019}$ & $.044\!_{\pm\!.036}$ \\
Atelectasis & $.000\!_{\pm\!.000}$ & $.000\!_{\pm\!.000}$ & $.000\!_{\pm\!.000}$ & $.000\!_{\pm\!.001}$ & $.000\!_{\pm\!.000}$ & $.000\!_{\pm\!.000}$ & $.007\!_{\pm\!.011}$ & $.004\!_{\pm\!.009}$ & $.004\!_{\pm\!.006}$ & $.016\!_{\pm\!.026}$ & $\mathbf{.022\!_{\pm\!.025}}$ \\
Cardiomegaly & $.000\!_{\pm\!.000}$ & $.000\!_{\pm\!.000}$ & $.000\!_{\pm\!.000}$ & $.000\!_{\pm\!.000}$ & $.000\!_{\pm\!.000}$ & $.000\!_{\pm\!.000}$ & $.007\!_{\pm\!.010}$ & $.001\!_{\pm\!.002}$ & $.012\!_{\pm\!.012}$ & $.003\!_{\pm\!.005}$ & $\mathbf{.020\!_{\pm\!.014}}$ \\
Consolidation & $.000\!_{\pm\!.000}$ & $.000\!_{\pm\!.000}$ & $.000\!_{\pm\!.000}$ & $.000\!_{\pm\!.000}$ & $.000\!_{\pm\!.000}$ & $.000\!_{\pm\!.000}$ & $.003\!_{\pm\!.007}$ & $.000\!_{\pm\!.000}$ & $.002\!_{\pm\!.002}$ & $\mathbf{.014\!_{\pm\!.018}}$ & $.012\!_{\pm\!.025}$ \\
Pneumonia & $.000\!_{\pm\!.000}$ & $.000\!_{\pm\!.000}$ & $.000\!_{\pm\!.000}$ & $.000\!_{\pm\!.000}$ & $.000\!_{\pm\!.000}$ & $.000\!_{\pm\!.000}$ & $.006\!_{\pm\!.009}$ & $.004\!_{\pm\!.012}$ & $.004\!_{\pm\!.012}$ & $\mathbf{.007\!_{\pm\!.008}}$ & $.007\!_{\pm\!.012}$ \\
Pneumothorax & $.000\!_{\pm\!.000}$ & $.000\!_{\pm\!.000}$ & $.000\!_{\pm\!.000}$ & $.000\!_{\pm\!.001}$ & $.000\!_{\pm\!.000}$ & $.000\!_{\pm\!.000}$ & $.000\!_{\pm\!.000}$ & $.002\!_{\pm\!.003}$ & $.003\!_{\pm\!.006}$ & $.004\!_{\pm\!.008}$ & $\mathbf{.007\!_{\pm\!.008}}$ \\
Subcut. Emphysema & $.000\!_{\pm\!.000}$ & $.000\!_{\pm\!.000}$ & $.000\!_{\pm\!.000}$ & $.000\!_{\pm\!.000}$ & $.000\!_{\pm\!.000}$ & $.000\!_{\pm\!.000}$ & $.000\!_{\pm\!.000}$ & $.000\!_{\pm\!.000}$ & $\mathbf{.004\!_{\pm\!.012}}$ & $.004\!_{\pm\!.012}$ & $.000\!_{\pm\!.000}$ \\
Calc. of the Aorta & $.000\!_{\pm\!.000}$ & $.000\!_{\pm\!.000}$ & $.000\!_{\pm\!.000}$ & $.000\!_{\pm\!.000}$ & $.000\!_{\pm\!.000}$ & $.000\!_{\pm\!.000}$ & $.000\!_{\pm\!.000}$ & $.000\!_{\pm\!.000}$ & $.000\!_{\pm\!.000}$ & $.000\!_{\pm\!.000}$ & $.000\!_{\pm\!.000}$ \\
Pneumomediastinum & $.000\!_{\pm\!.000}$ & $.000\!_{\pm\!.000}$ & $.000\!_{\pm\!.000}$ & $.000\!_{\pm\!.000}$ & $.000\!_{\pm\!.000}$ & $.000\!_{\pm\!.000}$ & $.000\!_{\pm\!.000}$ & $.000\!_{\pm\!.000}$ & $.000\!_{\pm\!.000}$ & $.000\!_{\pm\!.000}$ & $.000\!_{\pm\!.000}$ \\
Pneumoperitoneum & $.000\!_{\pm\!.000}$ & $.000\!_{\pm\!.000}$ & $.000\!_{\pm\!.000}$ & $.000\!_{\pm\!.000}$ & $.000\!_{\pm\!.000}$ & $.000\!_{\pm\!.000}$ & $.000\!_{\pm\!.000}$ & $.000\!_{\pm\!.000}$ & $.000\!_{\pm\!.000}$ & $.000\!_{\pm\!.000}$ & $.000\!_{\pm\!.000}$ \\
Tortuous Aorta & $.000\!_{\pm\!.000}$ & $.000\!_{\pm\!.000}$ & $.000\!_{\pm\!.000}$ & $.000\!_{\pm\!.000}$ & $.000\!_{\pm\!.000}$ & $.000\!_{\pm\!.000}$ & $.000\!_{\pm\!.000}$ & $.000\!_{\pm\!.000}$ & $.000\!_{\pm\!.000}$ & $.000\!_{\pm\!.000}$ & $.000\!_{\pm\!.000}$ \\
\midrule
\textbf{Macro Avg} & $.079\!_{\pm\!.007}$ & $.068\!_{\pm\!.000}$ & $.068\!_{\pm\!.000}$ & $.069\!_{\pm\!.002}$ & $.068\!_{\pm\!.000}$ & $.068\!_{\pm\!.000}$ & $.076\!_{\pm\!.003}$ & $.070\!_{\pm\!.003}$ & $.076\!_{\pm\!.004}$ & $.077\!_{\pm\!.005}$ & $\mathbf{.080\!_{\pm\!.006}}$ \\
\bottomrule
\end{tabular}%
}
\end{table*}

\section{Partition sensitivity and routing controls: full results}
\label{app:partition}

Table~\ref{tab:partition-full} reports all 12 feature-space metrics for the controls summarized in Table~\ref{tab:partition-routing}, and for learned versus fixed routing under minority guidance. Random reassignments are generated within the construction of Section~\ref{sec:method}: each map keeps the four experts, the reserved healthy expert and whole-label assignment, and fills the three pathology groups toward the same target loads as the submitted map; the map is fixed across training and evaluation. The $K=3$ and $K=5$ variants use two and four pathology groups plus the healthy expert. The literal head/medium/tail variant routes the official MIMIC-CXR-LT frequency strata~\cite{holste2022} to three experts, and the per-label variant uses one expert for each of the 19 labels. The two starved designs degrade mainly in label-macro FID, whose large across-label SD reflects the collapse of the rarest labels. For downstream recovery, the random maps are pooled over three maps $\times$ three classifier seeds; individually they recover $9$, $7$ and $5$ of $13$ classes.

\begin{table*}[!t]
\centering
\caption{Full feature-space metrics for the partition-sensitivity, routing and external-baseline controls on MIMIC-CXR-LT (GRASP-LowDim width unless stated). All-label $\pm$ is the SD over three generator seeds (submitted GRASP-LowDim and CCUA) or over the three random maps; other variants use generator seed 0 and are point estimates. Macro $\pm$ is the SD across the 19 label-wise values, as in Tables~\ref{tab:cxr-metrics} and~\ref{tab:dino-metrics}. CCUA is trained for 20k steps at per-GPU batch size 2 (half of GRASP's sample exposures; $\sim$28~h on 8 GPUs per seed, 2--4$\times$ GRASP's adapter training). The two +MG rows compare learned and fixed routing under minority guidance (three generator seeds each).}
\label{tab:partition-full}
\resizebox{\textwidth}{!}{%
\begin{tabular}{l c *{6}{c}}
\toprule
& & \multicolumn{3}{c}{\textbf{All labels}} & \multicolumn{3}{c}{\textbf{Macro Avg}} \\
\cmidrule(lr){3-5}\cmidrule(lr){6-8}
\textbf{Configuration} & \textbf{$K$} & \textbf{Cov.} $\uparrow$ & \textbf{IRS} $\uparrow$ & \textbf{FID} $\downarrow$ & \textbf{Cov.} $\uparrow$ & \textbf{IRS} $\uparrow$ & \textbf{FID} $\downarrow$ \\
\midrule
\multicolumn{8}{l}{\textit{DINOv2 features}} \\
Submitted partition (GRASP-LowDim) & 4  & $.009\!_{\pm\!.001}$ & $.169\!_{\pm\!.003}$ & $91.36\!_{\pm\!1.59}$  & $.116\!_{\pm\!.098}$ & $.376\!_{\pm\!.166}$ & $150.59\!_{\pm\!76.15}$ \\
Random reassignment (mean of 3)    & 4  & $.012\!_{\pm\!.000}$ & $.196\!_{\pm\!.001}$ & $84.22\!_{\pm\!4.11}$  & $.128\!_{\pm\!.097}$ & $.394\!_{\pm\!.150}$ & $139.63\!_{\pm\!66.06}$ \\
\quad Random map 0                 & 4  & $.012$ & $.194$ & $87.26$ & $.126\!_{\pm\!.093}$ & $.381\!_{\pm\!.138}$ & $142.91\!_{\pm\!78.86}$ \\
\quad Random map 1                 & 4  & $.012$ & $.195$ & $86.98$ & $.119\!_{\pm\!.082}$ & $.397\!_{\pm\!.156}$ & $142.46\!_{\pm\!75.37}$ \\
\quad Random map 2                 & 4  & $.011$ & $.198$ & $78.41$ & $.139\!_{\pm\!.128}$ & $.403\!_{\pm\!.165}$ & $133.51\!_{\pm\!62.54}$ \\
Fewer pathology groups             & 3  & $.012$ & $.189$ & $89.48$  & $.119\!_{\pm\!.078}$ & $.404\!_{\pm\!.167}$ & $149.65\!_{\pm\!72.57}$ \\
More pathology groups              & 5  & $.010$ & $.186$ & $94.16$  & $.110\!_{\pm\!.101}$ & $.421\!_{\pm\!.192}$ & $150.67\!_{\pm\!75.88}$ \\
Literal official head/medium/tail             & 3  & $.012$ & $.182$ & $80.61$  & $.071\!_{\pm\!.060}$ & $.388\!_{\pm\!.218}$ & $317.55\!_{\pm\!490.91}$ \\
One expert per label               & 19 & $.009$ & $.188$ & $106.59$ & $.035\!_{\pm\!.023}$ & $.384\!_{\pm\!.184}$ & $447.19\!_{\pm\!468.42}$ \\
Static-initialized learned gate    & 4  & $.010$ & $.187$ & $94.80$  & $.106\!_{\pm\!.100}$ & $.408\!_{\pm\!.170}$ & $149.96\!_{\pm\!74.13}$ \\
MoE-GRASP-LowDim + MG (learned) & 4 & $.012\!_{\pm\!.001}$ & $.218\!_{\pm\!.002}$ & $90.55\!_{\pm\!4.17}$ & $.112\!_{\pm\!.096}$ & $.412\!_{\pm\!.173}$ & $151.35\!_{\pm\!65.46}$ \\
GRASP-LowDim + MG (fixed)       & 4 & $.014\!_{\pm\!.001}$ & $.200\!_{\pm\!.004}$ & $83.99\!_{\pm\!1.58}$ & $.140\!_{\pm\!.107}$ & $.429\!_{\pm\!.191}$ & $142.04\!_{\pm\!75.33}$ \\
CCUA~\cite{ccua2025}           & -- & $.010\!_{\pm\!.003}$ & $.100\!_{\pm\!.007}$ & $117.72\!_{\pm\!39.49}$ & $.107\!_{\pm\!.083}$ & $.247\!_{\pm\!.099}$ & $157.50\!_{\pm\!26.43}$ \\
\midrule
\multicolumn{8}{l}{\textit{CXR features}} \\
Submitted partition (GRASP-LowDim) & 4  & $.942\!_{\pm\!.001}$ & $.503\!_{\pm\!.003}$ & $.0139\!_{\pm\!.0009}$ & $.924\!_{\pm\!.086}$ & $.555\!_{\pm\!.125}$ & $.0368\!_{\pm\!.0465}$ \\
Random reassignment (mean of 3)    & 4  & $.953\!_{\pm\!.003}$ & $.687\!_{\pm\!.006}$ & $.0119\!_{\pm\!.0002}$ & $.926\!_{\pm\!.094}$ & $.617\!_{\pm\!.099}$ & $.0368\!_{\pm\!.0386}$ \\
\quad Random map 0                 & 4  & $.955$ & $.695$ & $.0116$ & $.934\!_{\pm\!.088}$ & $.601\!_{\pm\!.126}$ & $.0354\!_{\pm\!.0432}$ \\
\quad Random map 1                 & 4  & $.948$ & $.685$ & $.0122$ & $.921\!_{\pm\!.088}$ & $.625\!_{\pm\!.122}$ & $.0376\!_{\pm\!.0372}$ \\
\quad Random map 2                 & 4  & $.954$ & $.680$ & $.0118$ & $.923\!_{\pm\!.114}$ & $.626\!_{\pm\!.134}$ & $.0375\!_{\pm\!.0373}$ \\
Fewer pathology groups             & 3  & $.952$ & $.660$ & $.0139$ & $.922\!_{\pm\!.094}$ & $.665\!_{\pm\!.157}$ & $.0406\!_{\pm\!.0379}$ \\
More pathology groups              & 5  & $.941$ & $.656$ & $.0138$ & $.920\!_{\pm\!.108}$ & $.580\!_{\pm\!.137}$ & $.0393\!_{\pm\!.0384}$ \\
Literal official head/medium/tail             & 3  & $.944$ & $.624$ & $.0181$ & $.887\!_{\pm\!.121}$ & $.503\!_{\pm\!.165}$ & $.0754\!_{\pm\!.0788}$ \\
One expert per label               & 19 & $.956$ & $.678$ & $.0105$ & $.903\!_{\pm\!.119}$ & $.540\!_{\pm\!.170}$ & $.0496\!_{\pm\!.0683}$ \\
Static-initialized learned gate    & 4  & $.948$ & $.675$ & $.0133$ & $.938\!_{\pm\!.085}$ & $.652\!_{\pm\!.160}$ & $.0383\!_{\pm\!.0464}$ \\
MoE-GRASP-LowDim + MG (learned) & 4 & $.941\!_{\pm\!.005}$ & $.632\!_{\pm\!.019}$ & $.0118\!_{\pm\!.0009}$ & $.920\!_{\pm\!.096}$ & $.604\!_{\pm\!.166}$ & $.0512\!_{\pm\!.0622}$ \\
GRASP-LowDim + MG (fixed)       & 4 & $.949\!_{\pm\!.000}$ & $.679\!_{\pm\!.002}$ & $.0135\!_{\pm\!.0008}$ & $.921\!_{\pm\!.093}$ & $.614\!_{\pm\!.122}$ & $.0392\!_{\pm\!.0478}$ \\
CCUA~\cite{ccua2025}           & -- & $.636\!_{\pm\!.031}$ & $.126\!_{\pm\!.006}$ & $.0671\!_{\pm\!.0035}$ & $.673\!_{\pm\!.153}$ & $.202\!_{\pm\!.178}$ & $.1457\!_{\pm\!.1144}$ \\
\bottomrule
\end{tabular}%
}
\end{table*}

\section{Compute Resources}
\label{app:compute}

All experiments were conducted on a shared academic research cluster using NVIDIA H100 GPUs with 80GB memory per GPU. Training and sampling were implemented in PyTorch and used mixed precision. We report compute in H100 GPU-hours, computed as wall-clock runtime multiplied by the number of GPUs.

Each generative model was fine-tuned for 10,000 steps (CCUA: 20,000) with a batch size of 8 on 8 H100 GPUs. A full fine-tuning run required approximately 30 wall-clock hours, corresponding to about 240 H100 GPU-hours per seed. Adapter-based runs, including GRASP variants, required approximately 7-12 wall-clock hours, corresponding to about 56-96 H100 GPU-hours per seed. For each model and seed, we generated 50,000 images at $1024 \times 1024$ resolution. This required approximately 12 wall-clock hours on 8 H100 GPUs, plus additional feature extraction and metric computation.

The primary GRASP configuration, evaluated over three generative seeds, requires approximately 550-650 H100 GPU-hours including training, image generation, and metric computation. Reproducing the primary comparison against the strongest baseline requires approximately 1,100-1,300 H100 GPU-hours.

Reproducing the main MIMIC-CXR-LT generation results, including full fine-tuning baselines, MoCLE, sequential adapters, GRASP variants, learned-routing variants, and minority-guided sampling evaluations, requires approximately 5,500-6,500 H100 GPU-hours. Including the ImageNet-LT experiments increases the estimate to approximately 9,500-11,500 H100 GPU-hours.

The downstream DenseNet-121 evaluation uses three classifier seeds on each of three synthetic dataset seeds, yielding nine classifier runs per configuration. Including these downstream classifier runs, the ablation experiments, gradient-analysis runs, and qualitative sample generation, we estimate that reproducing all reported quantitative results requires approximately 11,000-14,000 H100 GPU-hours.

The additional controls (from Slurm accounting) required approximately 4,100 GPU-hours: about 2,300 for the partition-sensitivity and static-initialized learned-gate runs (eight GRASP-LowDim trainings of about 150 GPU-hours each, one generator seed, plus generation, evaluation and classifiers), about 1,000 for CCUA (three seeds of 20k steps, about 28 wall-clock hours on 8 GPUs per seed, plus generation and classifiers), and about 840 for the fixed-router minority-guidance regenerations. Reproducing all reported quantitative results therefore requires approximately 15,000-18,000 GPU-hours.

Including exploratory development, debugging runs, failed configurations, preliminary hyperparameter choices, and intermediate ablations, we estimate the total compute used over the course of the project to be approximately 22,000-32,000 GPU-hours.

\newpage

\end{document}